\documentclass[conference]{IEEEtran}
\IEEEoverridecommandlockouts

\usepackage{cite}
\usepackage{amsmath,amssymb,amsfonts}
\usepackage{algorithmic}
\usepackage{graphicx}
\usepackage{textcomp}
\def\BibTeX{{\rm B\kern-.05em{\sc i\kern-.025em b}\kern-.08em
    T\kern-.1667em\lower.7ex\hbox{E}\kern-.125emX}}

  \usepackage{booktabs}
  \usepackage{hyperref}

\usepackage{graphicx}
\usepackage{subcaption}
\usepackage{enumitem}
\usepackage{float}

\usepackage[acronym]{glossaries}
\usepackage[dvipsnames]{xcolor}

\makeglossaries

\newacronym{apus}{APUs}{Application Processing Units}
\newacronym{tpu}{TPU}{Tensor Processing Unit}
\newacronym{sm}{SM}{sign-magnitude}
\newacronym{tc}{TC}{two's complement}
\newacronym{tcs}{TCS}{two's complement symmetric}
\newacronym{eda}{EDA}{Electronic Design Automation}
\newacronym{mig}{MIG}{Majority-Inverter-Graph}
\newacronym{vlsi}{VLSI}{Very-Large-Scale Integration}
\newacronym{mbe}{MBE}{Modified Booth Encoding}
\newacronym{gdi}{GDI}{Gate-Diffusion Input}
\newacronym{sme}{SME}{sign-magnitude extended}
\newacronym{ai}{AI}{Artificial Intelligence}
\newacronym{rtl}{RTL}{register-transfer level}
\newacronym{mac}{MAC}{multiply-accumulate}
\newacronym{aig}{AIG}{Add-Inverter Graph}
\newacronym{ieee}{IEEE}{Institute of Electrical and Electronics Engineers}
\newacronym{swact}{SwAct}{switching activity}
\newacronym{mlp}{MLP}{multi-layer perceptron}
\newacronym{iid}{\textit{iid}}{independent and identically distributed}
\newacronym{qor}{QoR}{Quality of Results}
\newacronym{epfl}{EPFL}{École Polytechnique Fédérale de Lausanne}
\newacronym{dse}{DSE}{design space exploration}
    
\begin{document}

\title{Explicit Sign-Magnitude Encoders \\ Enable  Power-Efficient Multipliers}




%
%
%
\author{
\IEEEauthorblockN{Felix Arnold, Maxence Bouvier, Ryan Amaudruz, Renzo Andri, Lukas Cavigelli}\\
\IEEEauthorblockA{Computing Systems Laboratory, Zurich Research Center, Huawei Technologies, Switzerland}
}

\markboth{34th International Workshop
on Logic \& Synthesis, June 12–13, 2025, University of Verona, Verona, Italy}%
{Shell \MakeLowercase{\textit{et al.}}: Bare Demo of IEEEtran.cls for IEEE Journals}
%



\maketitle

\pagestyle{plain}
\thispagestyle{plain}

\begin{abstract}

\\

This work presents a method to maximize power-efficiency of fixed point multiplier units by decomposing them into sub-components.
First, an encoder block converts the operands from a two's complement to a sign magnitude representation, followed by a multiplier module which performs the compute operation and outputs the resulting value in the original format.
This allows to leverage the power-efficiency of the \acrlong{sm} encoding for the multiplication.
To ensure the computing format is not altered, those two components are synthesized and optimized separately.
Our method leads to significant power savings for input values centered around zero, as commonly encountered in AI workloads.
Under a realistic input stream with values normally distributed with a standard deviation of 3.0, post-synthesis simulations of the 4-bit multiplier design show up to 12.9\% lower switching activity compared to synthesis without decomposition.
Those gains are achieved while ensuring compliance into any production-ready system as the overall circuit stays logic-equivalent.
With the compliance lifted and a slightly smaller input range of -7 to +7, switching activity reductions can reach up to 33\%.
Additionally, we demonstrate that synthesis optimization methods based on switching-activity-driven design space exploration can yield a further 5-10\% improvement in power-efficiency compared to a power agnostic approach.

\end{abstract}


%
\IEEEpeerreviewmaketitle

\begin{figure}[b]
\centering
\includegraphics[width=3.5in]{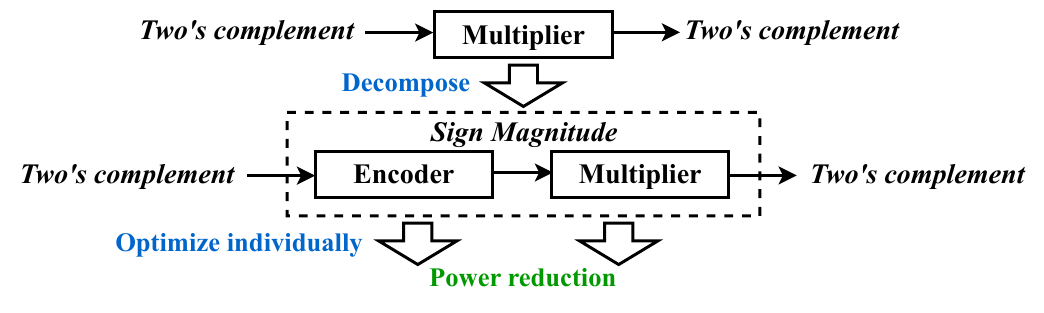}
\caption{Decomposition and optimization process utilizing a \acrlong{tc} to \acrlong{sm} encoder}
\label{fig:decomposition}
\end{figure}

\section{Introduction}

\IEEEPARstart {R}{ecent} advancements in specialized hardware for high-performance computing, such as the \gls{tpu} of Google, NVIDIA Tensor Cores, or Huawei's Ascend \gls{ai} processors, have enabled significant acceleration of deep learning algorithms, with major breakthroughs in many fields (physics, medicine, computer vision, reasoning, etc.).
At the core of these \gls{apus}, a large number of distributed processing elements are deployed to enable large scale arithmetic operations in parallel.
This massive number of computing elements along with a carefully designed memory hierarchy offers outstanding computing throughput, reaching tens of 4-bit PetaFLOPS ($10^{15}$ 4-bit operations per second) on-chip \cite{blackwell_tech_brief}.
Among these, multipliers play a critical role, impacting directly the efficiency of the arithmetic operations dominant in AI workloads.
Consequently, optimizing multipliers has become an important challenge in modern digital design, highlighting the need for high-quality combinational digital circuits to achieve optimal performance under a stringent power and area budget.

This work investigates a holistic approach to multiplier optimization through signed fixed point data formats, with the goal of enhancing the area and energy efficiency of the overall system.
After introducing metrics and data representations, we present a model to derive \gls{swact} from a cell-level post-synthesis netlist simulation.
In this work, the total transistor count is used as a proxy to the circuit area, while the power/energy consumption is assessed via our proposed \gls{swact} model.
\Gls{sm} number formats are generally more energy-efficient  \cite{waeijen2018} than the commonly used \gls{tc} for a wide range of workloads.
We show that explicitly breaking down the multiplier into an encoder (\gls{tc} to \gls{sm} conversion) and \gls{sm} multiplier can significantly increase the energy efficiency -- while staying functionally identical.
Conversely, we observe that \gls{eda} logic optimization methods fail to find such solutions, leaving substantial area and power minimization unexploited.
We report the area and power usage for different \gls{tc}-to-\gls{sm}-to-\gls{tc} multiplier configurations and demonstrate that enforcing the intermediate \acrlong{sm} representation can significantly reduce both transistor count and power consumption of a multiplier unit. It is important to note that not all configurations in this work are functionally identical; i.e., the effective input range is slightly varied.
We found that \gls{eda} tools such as Yosys \cite{yosys} do not put high efforts on optimizing individual combinational circuits.
However, highly-optimized combinational elements are essential to our work. Therefore, we introduce a synthesis optimization process that improves the synthesis \gls{qor}.
Our optimization method, inspired by \cite{arnold2025latebreakingresultsart}, builds upon the framework for random exploration of \gls{mig}-based minimization proposed by \cite{dac24_mt}, incorporating a multi-criteria selection process to better guide the successive steps.

\section{State-of-the-Art}

Historically, much of the work on advancing multiplier design has focused on reducing transistor count, power consumption, and timing.
Early innovations, such as the Wallace tree method \cite{wallace1964} and Booth encoding \cite{booth1951}, established efficient partial product reduction techniques that continue to influence current designs.
\gls{mbe} schemes have further been developed including high-performance versions \cite{wen_chang_high_speed_booth}.
In parallel, power-saving strategies, including data-driven signal gating \cite{honarmand2006} and the Spurious Power Suppression Technique \cite{chen2007}, address dynamic power dissipation by minimizing unnecessary \gls{swact}.
Comparative studies \cite{tang2011} have highlighted that hybrid logic styles, combined with approximate computing methods \cite{mohammad2018}, offer promising trade-offs between precision and power-efficiency.
Finally, deep learning methods \cite{pasandi2020deep} \cite{xue2025domac} and reinforcement learning methods \cite{wang2024hierarchical} \cite{feng2024gomarl} \cite{zuo2023rl} have also been used to optimize multipliers.

Furthermore, data format plays a vital role in modulating \gls{swact}.
Although \acrlong{tc}  is the predominant representation used to encode fixed point signed numbers, its inherent carry propagation can increase switching activities and dynamic power \cite{SwitchinginmultipliersMsc}.
In contrast, sign-magnitude allows to reduce the number of switches when the input data toggles close to zero, thus reducing power consumption \cite{waeijen2018}.
Specialized floating point formats such as bfloat16 \cite{bfloat16_web} and HiFloat8 \cite{hifloat8} have been proposed and aim at optimizing the trade-off between precision and efficiency, considering both power and area, for \gls{ai} workloads.
While HiFloat8 internally employs the sign-magnitude encoding, it is applied exclusively to the exponent field.
For years, the bit-width of weights and activations in deep learning models — particularly during inference — has been pushed lower to alleviate computational and memory constraints, and four-bit quantization has recently been gaining traction for Large Language Models\cite{dettmers2023case}\cite{ashkboos2024quarot}.
The above innovations highlight the growing trend towards low-precision, energy-efficient arithmetic in \gls{vlsi} design.

\section{Data and Operations}
\subsection{Data Representations}
There are numerous ways to represent numbers with binary digits.
Commonly used representations can be classified into floating or fixed point formats, with further distinctions based on their ability to represent unsigned or signed values.
Some less common, but noteworthy recent representation include specialized data types for \gls{ai} workloads such as brain floating point bfloat16 \cite{bfloat16_web} or HiFloat8 \cite{hifloat8}.
In this work, we focus on four signed fixed point integer representations: \acrlong{tc}, \gls{tcs}, \acrlong{sm}, and \gls{sme} as defined in Table~\ref{table:three_representations}.
The \acrlong{tcs} and \acrlong{sm} representations cannot represent the full value space, as we exclude the most negative number 
(e.g., $-4$ in a 3-bit configuration). 
With the \gls{sme}, this shortcoming is fixed by re-adding the most negative number. By choice, we do not incorporate a negative zero in any format.

\begin{table*}[!ht] 
  \centering 
  \begin{tabular}{lccccccccc} 
  \toprule
  Value & --4 & --3 & --2 & --1 & 0 & 1 & 2 & 3 & Illegal\\
  \midrule
  Two's Complement (TC)& 100 & 101 & 110 & 111 & 000 & 001 & 010 & 011 & N/A\\
  Two's Compl. Symmetric (TCS) & N/A & 101 & 110 & 111 & 000 & 001 & 010 & 011 & 100\\
  Sign-Magnitude (SM) & N/A & 111 & 110 & 101 & 000 & 001 & 010 & 011 & 100 \\
  Sign-Magnitude Extended (SME) & 100 & 111 & 110 & 101 & 000 & 001 & 010 & 011 & N/A\\
  \bottomrule
  \end{tabular}
  \caption{Four possible representations of signed 3-bit numbers used in this work}
  \label{table:three_representations}
\end{table*}

\subsection{Encoder and Multiplier Module Implementations}

In this work, we focus exclusively on 2-input multipliers that receive 4-bit inputs and produce 8-bit outputs.
For the implementation of the different variants, we designed several \gls{rtl} descriptions of the 4-bit to 4-bit encoders and 2\texttimes 4-bit to 8-bit multiplier modules.
All functional blocks are defined in Verilog HDL language and were verified over the entire valid input space.

\subsubsection{Encoders}
An encoder performs the binary mapping between one representation to another.
For values represented with 4 bits, the encoder has a 4-bit input and a 4-bit output.
To cover all the possible two's complement to sign-magnitude encoding pairs, we designed three different encoders: \gls{tc}$\rightarrow$\gls{sm}, \gls{tc}$\rightarrow$\gls{sme} and \gls{tcs}$\rightarrow$\gls{sm}.
The \gls{tcs}$\rightarrow$\gls{sm} and \gls{tc}$\rightarrow$\gls{sme} mappings are simple variations of each-other and efficiently implemented with the same RTL code.
As the value $-8$ cannot be represented with the \acrlong{sm} encoding, the \gls{tc}$\rightarrow$\gls{sm} design integrates a supplementary clipping logic to map the two's complement representation of $-8$ to the value $-7$ in sign-magnitude representation.

\subsubsection{Multipliers}
The multiplier module takes in two 4-bit input operands encoded in a specific representation (\gls{tc}, \gls{tcs}, \gls{sm}, or \gls{sme}), performs the multiplication operation, and always outputs the 8-bit result in the \gls{tc} representation.
We thus implement and synthesize the following input/output format combinations: \gls{tc}$\rightarrow$\gls{tc}, \gls{sme}$\rightarrow$\gls{tc}, and  \gls{sm}$\rightarrow$\gls{tc}.
For both \gls{tc}$\rightarrow$\gls{tc} and \gls{sme}$\rightarrow$\gls{tc} multipliers, the entire dynamic ranges are allowed in both input and output, from $-8$ to $+7$ and from $-128$ to $+127$ respectively.
However, with \gls{sm}$\rightarrow$\gls{tc}, the most negative value of $-8$ is not included in the input space.

The \gls{tc}$\rightarrow$\gls{tc} multiplier is defined by using the \textit{star operator} in Verilog, and synthesized as a preconceived Booth multiplier by Yosys \cite{yosys}.
The \gls{sm}$\rightarrow$\gls{tc} implementation is straightforward; the unsigned magnitude parts of the input operands are multiplied together and the sign is determined using an XOR operation between the sign bit of the operands.
As the magnitude is represented in 3\,bits, the internal operation is performed by a 3b\texttimes3b unsigned multiplier, resulting in a smaller unit compared to the \gls{tc}$\rightarrow$\gls{tc} multiplier.
Finally, the result is converted to two's complement representation by inverting all the individual bits and adding 1  if the result is negative.

For the \gls{sme}$\rightarrow$\gls{tc} multiplier, we use the \gls{sm}$\rightarrow$\gls{tc} implementation as a starting point.
To extend the input space with the missing $-8$ value, one might consider replacing the 3-bit unsigned multiplier with a 4-bit version.
However, we observed that the alternative approach of replacing the operand by $4$ in case of a -8 input and left-shifting the result is more efficient.
E.g., $-8 \cdot 3$ would result in a magnitude calculation of $4 \cdot 3 = 12$ which would be left-shifted to $24$, and the final result (sign included) would then be $-24$.
Thus, internally, our implementation of the \gls{sme}$\rightarrow$\gls{tc} multiplier still multiplies using 3-bit values.

\subsection{Data Distribution}
\begin{figure*}[htbp]
\centering
\includegraphics[width=5.5in]{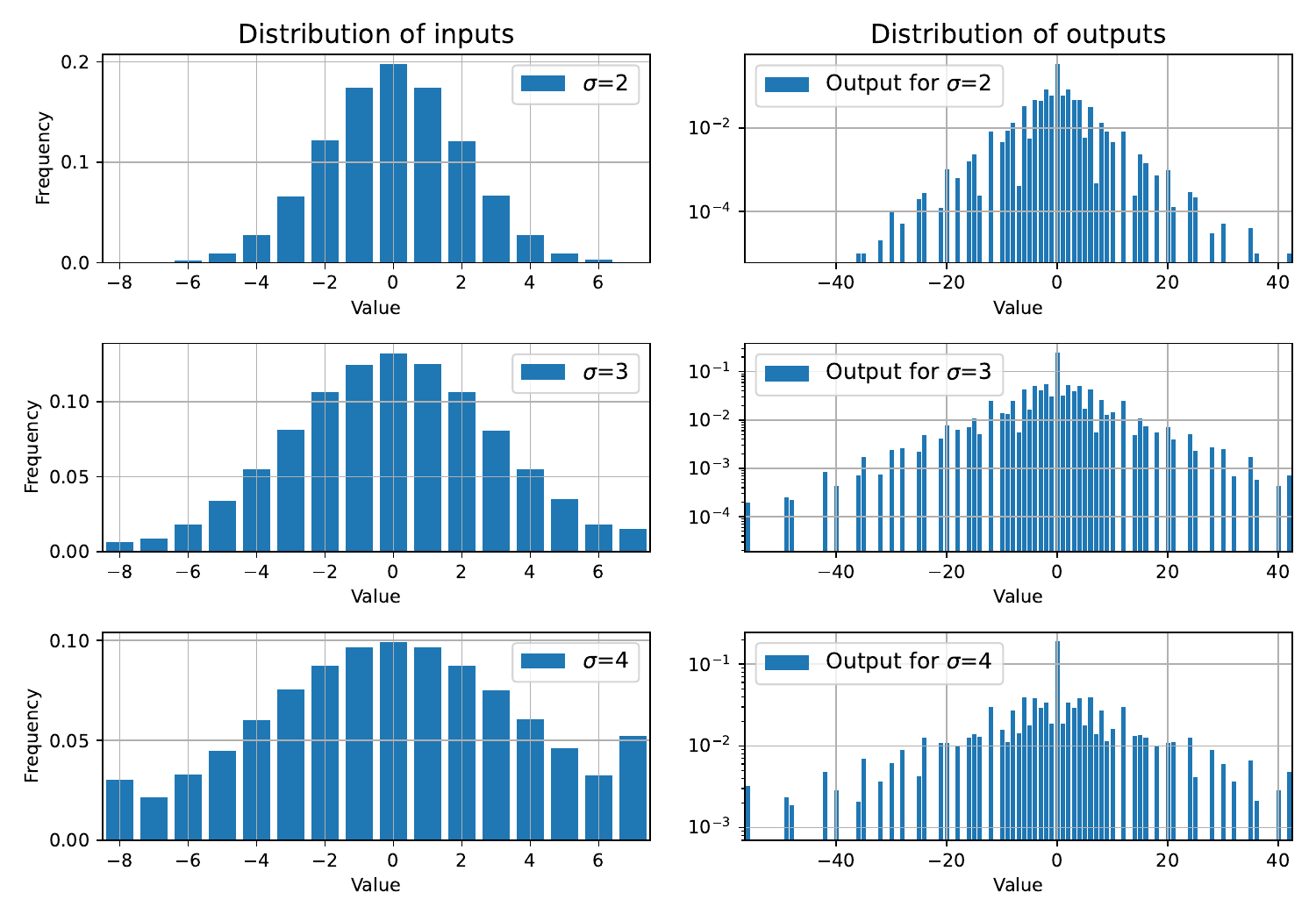}
\caption{Distribution of input and output values for a 4-bit two's complement multiplication}
\label{fig_distribution}
\end{figure*}

The distribution of the values fed to the multipliers largely depend on the workload.
In machine learning applications, particularly in large language models, weight distributions often approximate a normal distribution with standard deviation $\sigma$ and mean $\mu$ \cite{bfl}.
Therefore, for post-synthesis simulations, we sample input operands that are \gls{iid} and drawn from normal distributions centered around 0 with varying $\sigma$.
If a sampled value exceeds the range of the allowed representation, the value is clipped to the closest value in the range.
Histograms for different standard deviations are shown in Fig. \ref{fig_distribution}.
A standard deviation of $\sigma=3.0$ minimizes clipping while still utilizing the full representable range, aligning well with real-workloads, such as those in \cite{quantized_weights_supporting_material}. In practice, the actual value for $\sigma$ depends on several factors, including application, neural network architecture, training procedure, and quantization method.

\section{Methods}

\subsection{Quality Metrics}

To assess the synthesis \gls{qor}, we extract and capture the following metrics:

\begin{itemize}
\item \textbf{Number of cells}: The number of technology cells, as reported by YOSYS \cite{yosys} after a technology-independent \texttt{abc} step. Cells are standard logic cells such as AND, OR, NOT, XOR, etc..
\item \textbf{Number of transistors}: Cell area (or complexity) is usually simplified by giving the cell size equivalent in NAND count, i.e. the area of the cell divided by the area of a NAND count. This can be extrapolated to number of transistors as a NAND cells are associated with a fixed number of transistors. This NAND count is technology dependent \cite{Kaeslin14}, but YOSYS proposes a default approximate model of the number of transistors of each cell, which we employ to obtain this metric.
\item \textbf{Depth}: The depth of the circuit is derived on cell level by taking the longest path, in terms of number of cells, between an input and output along the circuit graph of the synthesized circuit. As we do not back-annotate the post-synthesis circuits with timing information, the complexity (or delay) of the cell is not taken into account. As a result, the depth of a circuit is directly related but not necessarily proportional to its maximal operating frequency.
\item \textbf{Switching activity}: The \gls{swact} is evaluated through post-synthesis simulation. A \gls{swact} model weights the reported toggling on each wire by the number of transistors in each cell. This metric is directly related to power, assuming that dynamic power is dominant.
\end{itemize}

\subsection{Switching Activity Evaluation}

\begin{figure}[htbp]
\centering
\includegraphics[width=3.0in]{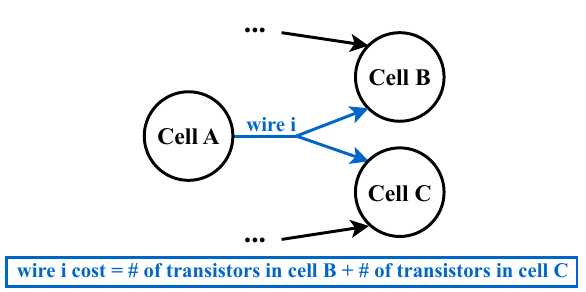}
\caption{Illustration of the proposed \acrlong{swact} model for a cell with a fanout of two downstream cells.}
\label{fig:swact_model}
\end{figure}

Compute-bound applications tend to achieve near-full utilization of a chip's compute elements over time.
Under such conditions, dynamic power is the dominant contributor and leakage plays a small role in the overall power-consumption of the system.
We are therefore interested in measuring and reducing the \gls{swact} within the logic of individual compute units.
To enable synthesis and \gls{swact} estimation of millions of multiplier designs in a reasonable time span, we propose the following model, as illustrated Fig. \ref{fig:swact_model}.

A post-synthesis \gls{rtl}-level simulation of the circuit mapped onto cells is executed.
At each clock cycle, a new pair of inputs (a new stimulus) is sampled from a predefined distribution and applied to the multiplier's input ports.
This is repeated a few thousands of times.
The resulting binary states of all individual inputs, outputs, and internal wires are logged for each cycle. 
Based on this data, the number of switches of each wire is evaluated by computing the difference of its states between two consecutive clock cycles.
To account for both the fan-out of each wire and the complexity of the fan-out cells, switches are assigned a cost factor. The cost of a wire is defined by the total number of transistors in all of its fan-out cells.
This captures the fact that a single wire state transition might result in multiple transistor switches.
The \gls{swact} of the circuit is derived by summing the cost-weighted switches of all individual wires, averaged over all clock cycles. Within this report, the \gls{swact} is always evaluated by simulating circuits for 10,000 clock cycles.
In this work, we denote \gls{swact} values with the arbitrary unit (a.u.) notation.

By avoiding timing back annotations and precise timing simulation of each circuit, this model allows for quick estimation of the circuit \gls{swact},  enabling a low cost power estimation.
We note that as the output ports do not have any cell attached to them, their fan-out weights are all zero.
As a result, output wires do not contribute to the \gls{swact} of the circuit.
Finally, the total \gls{swact} for a multiplier with two's complement input representation and two's complement output representation can be derived as: $s_{tot}=2 \cdot s_{enc} + s_{mult} \label{eq:s_mm}$, where $s_{enc}$ is the \gls{swact} of an encoder circuit and $s_{mult}$ is the \gls{swact} of the multiplier circuit.

\subsection{Optimization}
\label{sec:optimization}

One of the major issues when performing post-synthesis power estimation of a circuit is that the measurement depends on both the quality of the implementation of the design and its synthesis.
Unfortunately, we observed that most \gls{eda} tools synthesize multipliers by replacing a module inferred as a multiplier with preconceived and highly optimized macro cells, such as a radix-4 Booth-encoded implementation in the case of YOSYS \cite{yosys}.
Subsequently, optimization of these blocks by the synthesis tools is heavily biased and non-ideal.
In order to fully exploit custom multipliers, an equally high synthesis quality is required for all individual designs.
In light of this, we leverage the State-of-the-Art \gls{mig} and \gls{aig} based synthesis framework proposed by \cite{dac24_mt}.
This framework originally consists in a random-walk optimization where randomly selected \gls{mig} or \gls{aig}-based decompression and compression algorithms are successively applied in a step-wise fashion on a circuit, until no further improvement can be observed.

\begin{figure*}[htbp]
\centering
\includegraphics[width=7in]{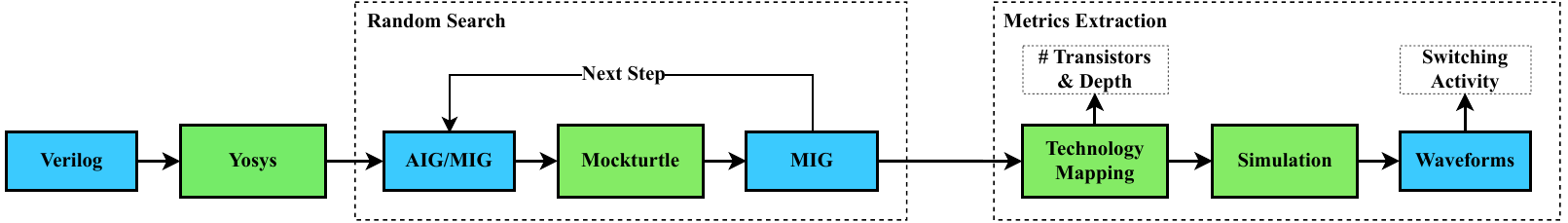}
\caption{High-level flow diagram}
\label{fig:high_level_flow}
\end{figure*}

We use the Mockturtle tool from the \acrshort{epfl} logic synthesis libraries \cite{EPFLLibraries} and reproduce their method with the following notable changes.
At each step, one recipe out of the thirty possible scripts (20 decompression script configurations and 10 compression ones) is randomly selected with equal probability and applied to the circuit.
When a compression script is chosen, the same script is applied three times in a row to ensure an appropriate balance between compression and decompression scripts.
This process is repeated for a fixed number of steps, which we call a chain of steps or an iteration.
Each iteration starts with the best circuit found in all previous chains.
Ultimately, the best circuit across all chains is selected.
This process is repeated across multiple independent runs, from which the overall best circuit is again chosen.
To decide which circuit to select, either the best number of transistors, the \gls{swact} or both are considered at the end of each chain.
Figure \ref{fig:high_level_flow} illustrates the high level data flow, excluding the chain and iteration mechanisms.
Overall, this iterative procedure mirrors the data flow described in \cite{arnold2025latebreakingresultsart}, where parallel sequences and progressive optimization converge to yield a globally optimal solution.

\begin{figure}[tbp]
\vspace{-0.25cm}
\centering
\subfloat[TC$\rightarrow$TC multiplier circuit]{\includegraphics[width=3.7in]{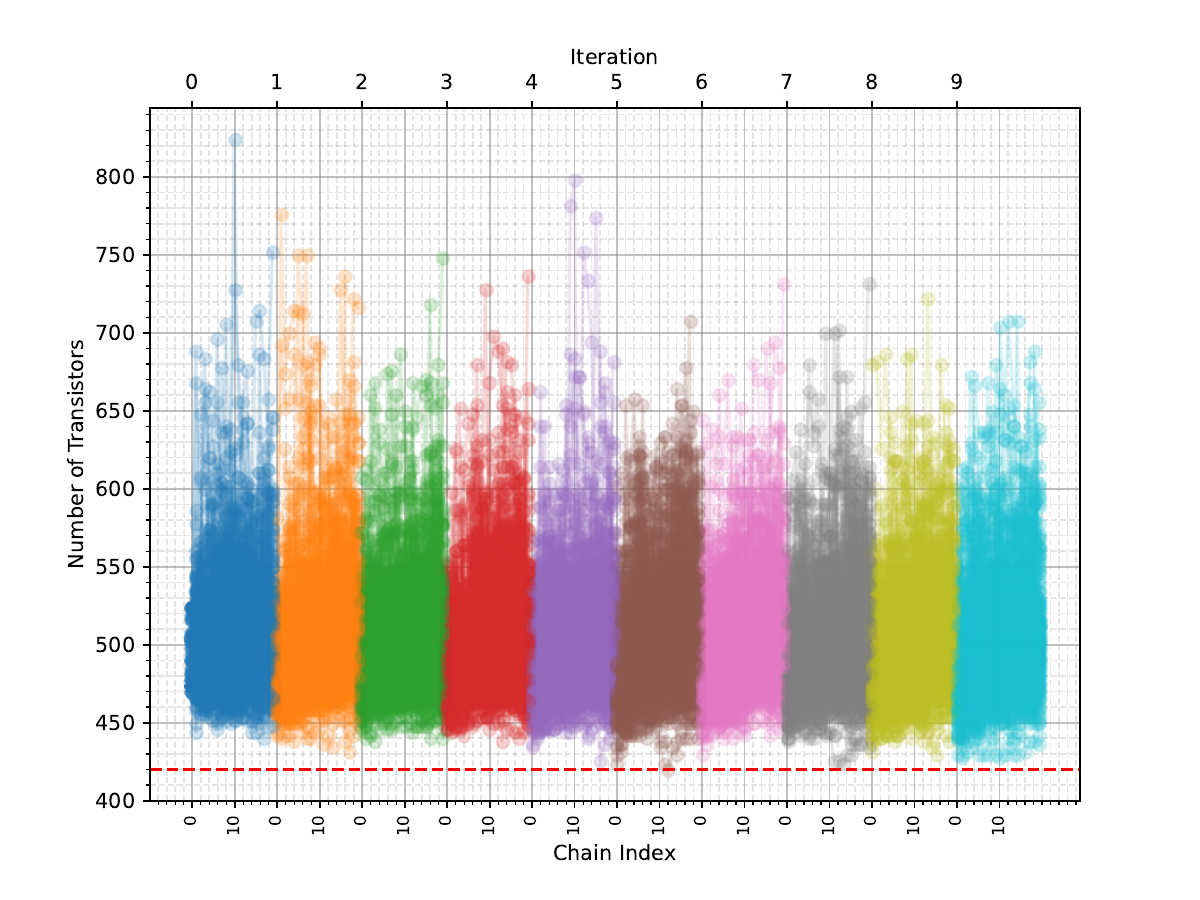}%
\label{fig:tc_to_tc}}
\hfil
\subfloat[SME$\rightarrow$TC multiplier circuit]{\includegraphics[width=3.7in]{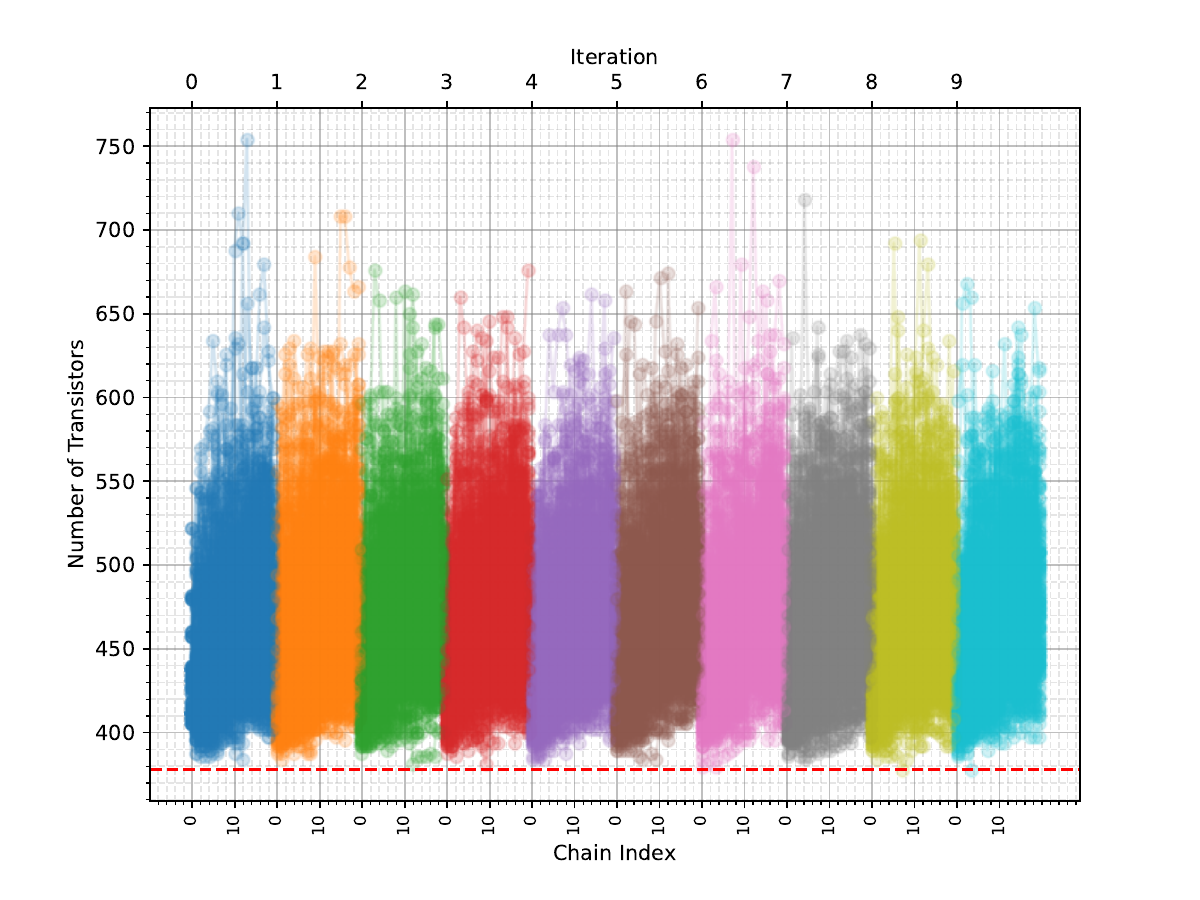}%
\label{fig_second_case}}
\caption{Optimization traces for 200 runs overlaid}
\label{fig:optim_traces}
\end{figure}

To conduct a fair comparison, all encoder and multiplier circuits presented in the Results section were optimized with the same level of effort.
The optimization process, as described above, is performed with 200 concurrent runs, each with 10 iterations of 20 steps, i.e. a total of 200 runs of 200 steps. 
The transistor count is the metric used to select the best circuit at each iteration within a run, while the best \gls{swact} (for $\sigma=3$) determines the final circuit selected across all runs.
The entire \gls{dse} process for two multiplier variants is depicted in \ref{fig:optim_traces}.

Finally, each of the encoder and multiplier blocks is optimized with the above flow to get to a single circuit which is then used to report the values presented in the Results section.

\section{Experiments}
\subsection{Configurations}

Several meaningful encoder-multiplier configurations are evaluated.
As previously stated, circuits with input and output compatible with the two's complement present the advantages of:
a) seamless integration in production-ready systems where signed integers are usually represented using the two's complement format; b) an \textit{add} operation usually follows the \textit{multiply} operation within a MAC operators, and the two's complement is very efficient for additions of signed numbers.
Nevertheless, for completeness, we also include the sign-magnitude input to sign-magnitude output variant.
Table \ref{tab:results_swact} lists the evaluated hardware configurations which are described in further detail below:\\
\begin{itemize}
    \item \textbf{Configuration A (baseline):} The basic 4-bit input and 8-bit output two's complement multiplier.

    \item \textbf{Configuration B:} Numerically equivalent to Configuration A, but the logic is decomposed into separate encoder and multiplier components to perform the multiplication in \gls{sme} format. Each sub-component is synthesized individually.

    \item \textbf{Configuration C:} Similarly to \textbf{B} the input is in two's complement format.
However, the most negative number (-8) is clipped to (-7) within the encoder module.
While the input and output ranges are the same as in \textbf{A} and \textbf{B}, this configuration is not numerically equivalent.

    \item \textbf{Configuration D:} Here, the most negative input value is not available.
Input and output are still expected in two's complement format, but with a symmetric input range.
Note that this configurations could be used with minimal overhead for neural network, e.g., by quantizing the weights into the reduced $-7$ to $7$ range, and dynamically clipping the activation values to $-7$.

    \item \textbf{Configuration E:} In contrast to all other configurations, the input and output are expected to be in sign-magnitude format.
\textbf{E} would be particularly efficient in a system where weights are pre-stored in \gls{sm}, as it can lead to significant power reductions.

\end{itemize}

\subsection{Results}

\begin{figure}[htbp]
    \centering
    \begin{subfigure}[b]{0.48\textwidth}
        \centering
        \includegraphics[width=\textwidth]{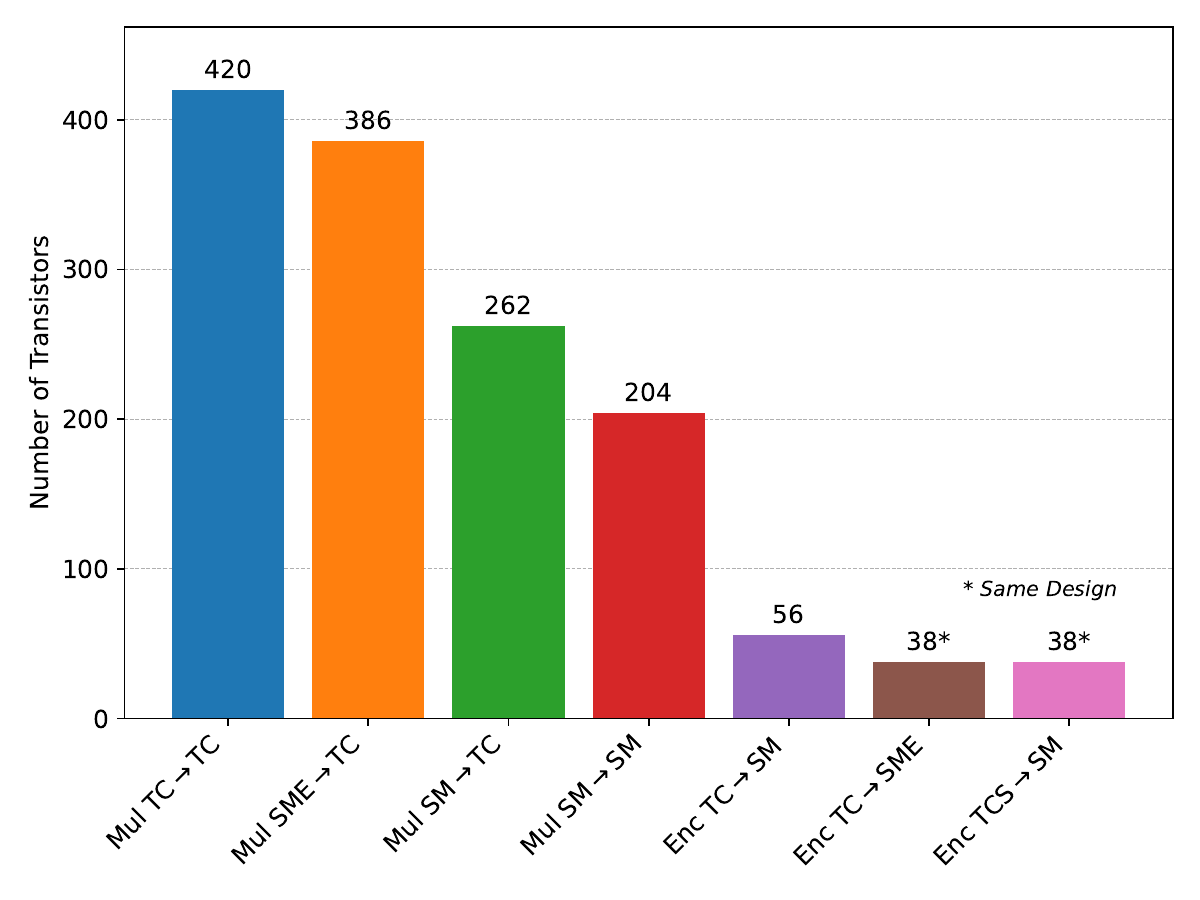}
        \caption{Transistors}
        \label{fig:sub1}
    \end{subfigure}
    \hfill
    \begin{subfigure}[b]{0.48\textwidth}
        \centering
        \includegraphics[width=\textwidth]{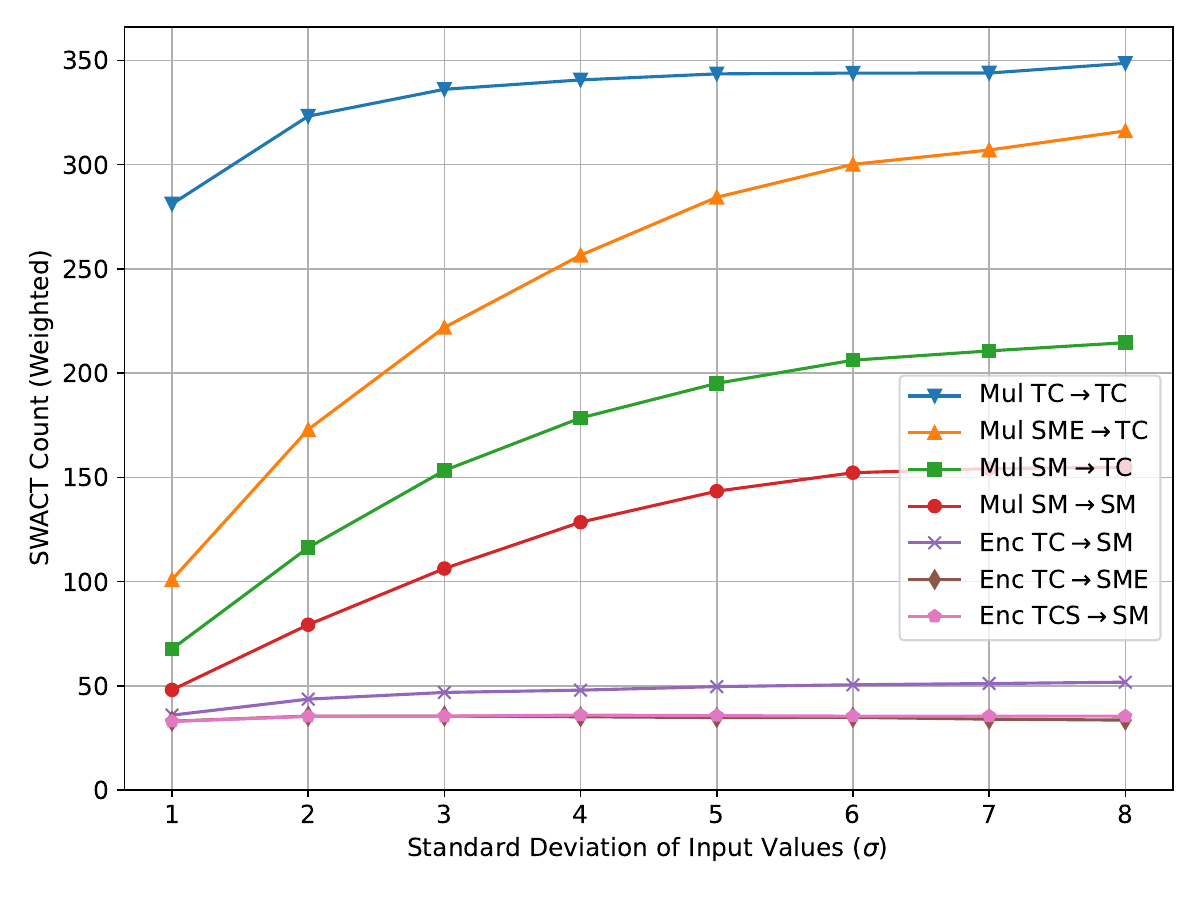}
        \caption{Switching Activity (a.u.)}
        \label{fig:sub2}
    \end{subfigure}
    \caption{Transistor count and switching activity for different multipliers and encoders}
    \label{fig:side_by_side}
\end{figure}

Fig. \ref{fig:side_by_side} depicts the results obtained when optimizing and simulating each sub-components individually.
Those results reveal that as the dynamic range is reduced, we see the area (number of transistors) and power (switching activity) diminish.
Fig. \ref{fig:side_by_side} a. depicts that the baseline \gls{tc}$\rightarrow$\gls{tc} (\textbf{A}) multiplier has the highest area with 420 transistors, followed closely by the \gls{sme}$\rightarrow$\gls{tc} (\textbf{B}) with 386 transistors.
The reduced dynamic range of the \gls{sm}-based multipliers (\textbf{C}, \textbf{D} and \textbf{E}) allows to reach lower areas of 262 and 204 transistors.
Fig. \ref{fig:side_by_side} b. shows the sign-magnitude variants have significantly lower \gls{swact} especially for normal input values distribution with lower $\sigma$.
Notably, even when adding the contribution of the encoders, they display substantially lower energy values.

A complete overview of the results for all configurations  is available in Table \ref{tab:results_swact}.
Considering \gls{swact}, all proposed configurations yield an advantage compared to the baseline.
For the circuit numerically equivalent to the baseline (configuration \textbf{B}) a $12.9\%$ improvement in \gls{swact} is obtained at $\sigma=3$ and is even more pronounced with $24.5\%$ at $\sigma=2$.
The lower-dynamic range but \gls{tc}-based configurations \textbf{C} and \textbf{D} reach a power reduction of up to $42.2\%$ at $\sigma=2$.
Finally, if we resort to  end-to-end \gls{sm} multiplication (\textbf{E}), a drastic \gls{swact} improvement of $75.5\%$ can be achieved for input distributions at $\sigma=2.0$.

We report the detailed transistor count and depth numbers in Table \ref{tab:results_trans_depth}.
For all decomposed configurations (\textbf{B}, \textbf{C}, \textbf{D}), the total depth is obtained by adding the depth of one encoder and one multiplier, while the total number of transistors is derived by adding the values for two encoders and one multiplier.
The results suggest that only the \gls{sme}-based configuration will induce a $10.0\%$ increase in area, while other configurations come with a reduction with respect to the baseline.
However, all decomposed configurations show a substantial increase in depth (from $16.7\%$ to $41.7\%$).
We expect that this could be partially alleviated by performing a supplementary synthesis of all components together; however, time constraints prevented us from running this experiment.
Additionally, it is important to note that the depth may not be directly proportional to the propagation delay, since not all cells have the same delay.

\begin{table*}[h]
    \centering
    \begin{tabular}{lll|cccc|cccc|cccc}
        \toprule
        \multicolumn{3}{c|}{\textbf{Hardware Configuration}} & \multicolumn{4}{c|}{\textbf{Switching Activity, $\sigma=2.0$}} & \multicolumn{4}{c|}{\textbf{Switching Activity, $\sigma=3.0$}} & \multicolumn{4}{c}{\textbf{Switching Activity, $\sigma=4.0$}} \\
        \cmidrule(lr){1-3} \cmidrule(lr){4-7} \cmidrule(lr){8-11} \cmidrule(lr){12-15}
         & \textbf{Encoder} & \textbf{Multiplier} & \textbf{Enc} & \textbf{Mul} & \textbf{Tot} & \textbf{$\Delta$\%} & \textbf{Enc} & \textbf{Mul} & \textbf{Tot} & \textbf{$\Delta$\%} & \textbf{Enc} & \textbf{Mul} & \textbf{Tot} & \textbf{$\Delta$\%} \\
        \toprule
         & $e$ & $m$ & ${s}_e$ & ${s}_m$ & ${s}_{tot}$ & -- & ${s}_e$ & ${s}_m$ & ${s}_{tot}$ & -- & ${s}_e$ & ${s}_m$ & ${s}_{tot}$ & -- \\
        \midrule
        A & None & TC$\rightarrow$TC & 0 & 323 & 323 & \textcolor{black}{0.0} & 0 & 336 & 336 & \textcolor{black}{0.0} & 0 & 341 & 341 & \textcolor{black}{0.0} \\
        B & TC$\rightarrow$SME & SME$\rightarrow$TC & 35 & 173 & 244 & \textcolor{Green}{-24.5} & 36 & 222 & 293 & \textcolor{Green}{-12.9} & 35 & 257 & 327 & \textcolor{Green}{-4.0} \\
        C & TC$\rightarrow$SM & SM$\rightarrow$TC & 44 & 116 & 204 & \textcolor{Green}{-37.0} & 47 & 153 & 247 & \textcolor{Green}{-26.5} & 48 & 178 & 274 & \textcolor{Green}{-19.5} \\
        D & TCS$\rightarrow$SM & SM$\rightarrow$TC & 35 & 116 & 187 & \textcolor{Green}{-42.2} & 35 & 153 & 224 & \textcolor{Green}{-33.3} & 36 & 178 & 250 & \textcolor{Green}{-26.5} \\
        E & None & SM$\rightarrow$SM & 0 & 79 & 79 & \textcolor{Green}{-75.5} & 0 & 106 & 106 & \textcolor{Green}{-68.4} & 0 & 128 & 128 & \textcolor{Green}{-62.3} \\
        \bottomrule
    \end{tabular}
    \caption{Switching activity for different encoder and multiplier settings, for $\sigma=2.0$, $\sigma=3.0$ and $\sigma=4.0$ (a.u.).}
    \label{tab:results_swact}
\end{table*}

\vspace{0.5cm}
\begin{table*}[h]
    \centering
    \begin{tabular}{lll|cccc|cccc}
        \toprule
        \multicolumn{3}{c|}{\textbf{Hardware Configuration}} & \multicolumn{4}{c|}{\textbf{Transistor Count}} & \multicolumn{4}{c}{\textbf{Depth}} \\
        \cmidrule(lr){1-3} \cmidrule(lr){4-7} \cmidrule(lr){8-11}
         & \textbf{Encoder} & \textbf{Multiplier} & \textbf{Enc} & \textbf{Mul} & \textbf{Tot} & \textbf{$\Delta$\%} & \textbf{Enc} & \textbf{Mul} & \textbf{Tot} & \textbf{$\Delta$\%} \\
        \toprule
         & $e$ & $m$ & ${t}_e$ & ${t}_m$ & ${t}_{tot}$ & -- & ${d}_e$ & ${d}_m$ & ${d}_{tot}$ & -- \\
        \midrule
        A & None & TC$\rightarrow$TC & 0 & 420 & 420 & \textcolor{black}{0.0} & 0 & 12 & 12 & \textcolor{black}{0.0} \\
        B & TC$\rightarrow$SME & SME$\rightarrow$TC & 38 & 386 & 462 & \textcolor{Mahogany}{10.0} & 3 & 14 & 17 & \textcolor{Mahogany}{41.7} \\
        C & TC$\rightarrow$SM & SM$\rightarrow$TC & 56 & 262 & 374 & \textcolor{Green}{-11.0} & 4 & 11 & 15 & \textcolor{Mahogany}{25.0} \\
        D & TCS$\rightarrow$SM & SM$\rightarrow$TC & 38 & 262 & 338 & \textcolor{Green}{-19.5} & 3 & 11 & 14 & \textcolor{Mahogany}{16.7} \\
        E & None & SM$\rightarrow$SM & 0 & 204 & 204 & \textcolor{Green}{-51.4} & 0 & 7 & 7 & \textcolor{Green}{-41.7} \\
        \bottomrule
    \end{tabular}
    \caption{Transistor count and depth for different encoder and multiplier settings}
    \label{tab:results_trans_depth}
\end{table*}

\section{Further Optimization Efforts}
\label{sec:further}
\begin{figure}[htbp]
\centering
\subfloat[No iterations]{\includegraphics[width=2.35in]{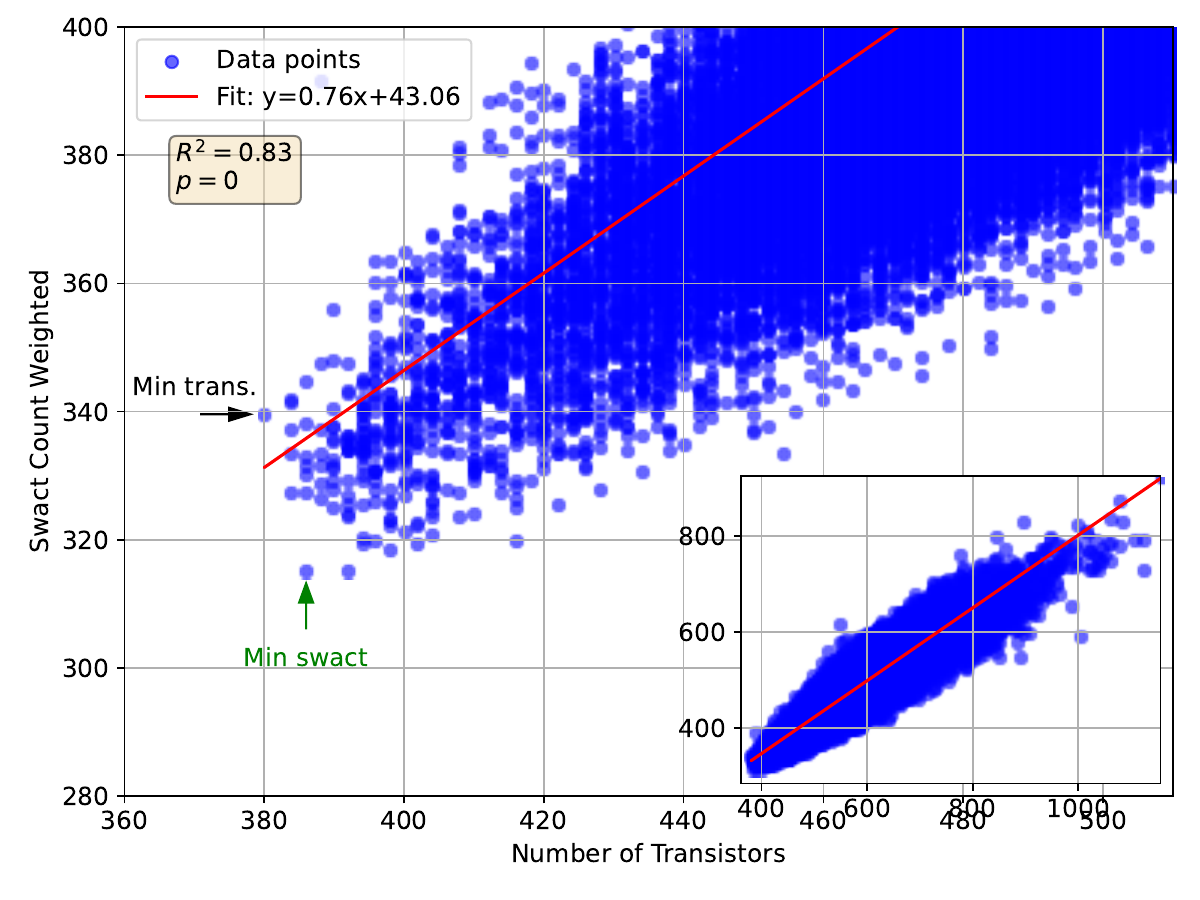}
\label{fig_first_caseA}
}\\[1ex]

\subfloat[Iterations with target transistors]{\includegraphics[width=2.35in]{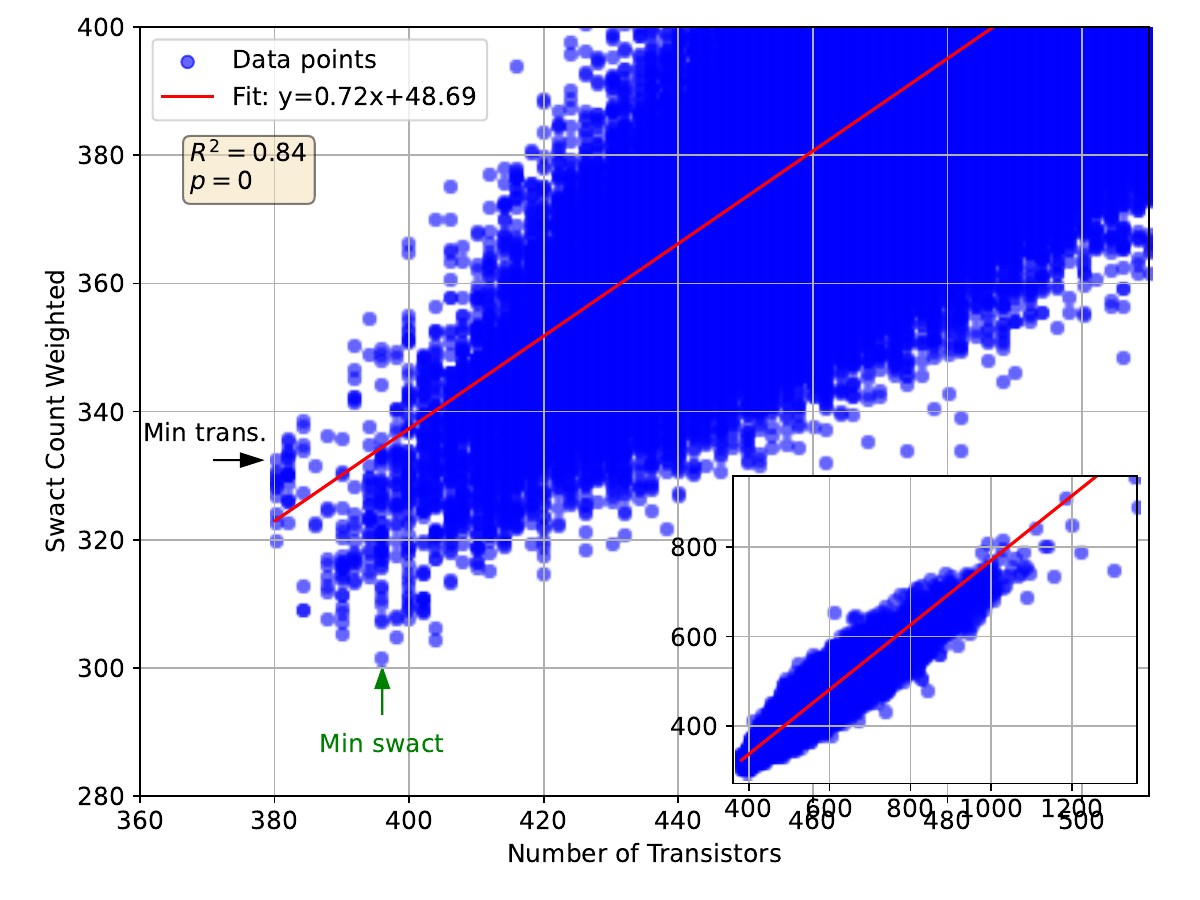}
\label{fig_second_caseB}
}\\[1ex]

\subfloat[Iterations with target switching activity]{\includegraphics[width=2.35in]{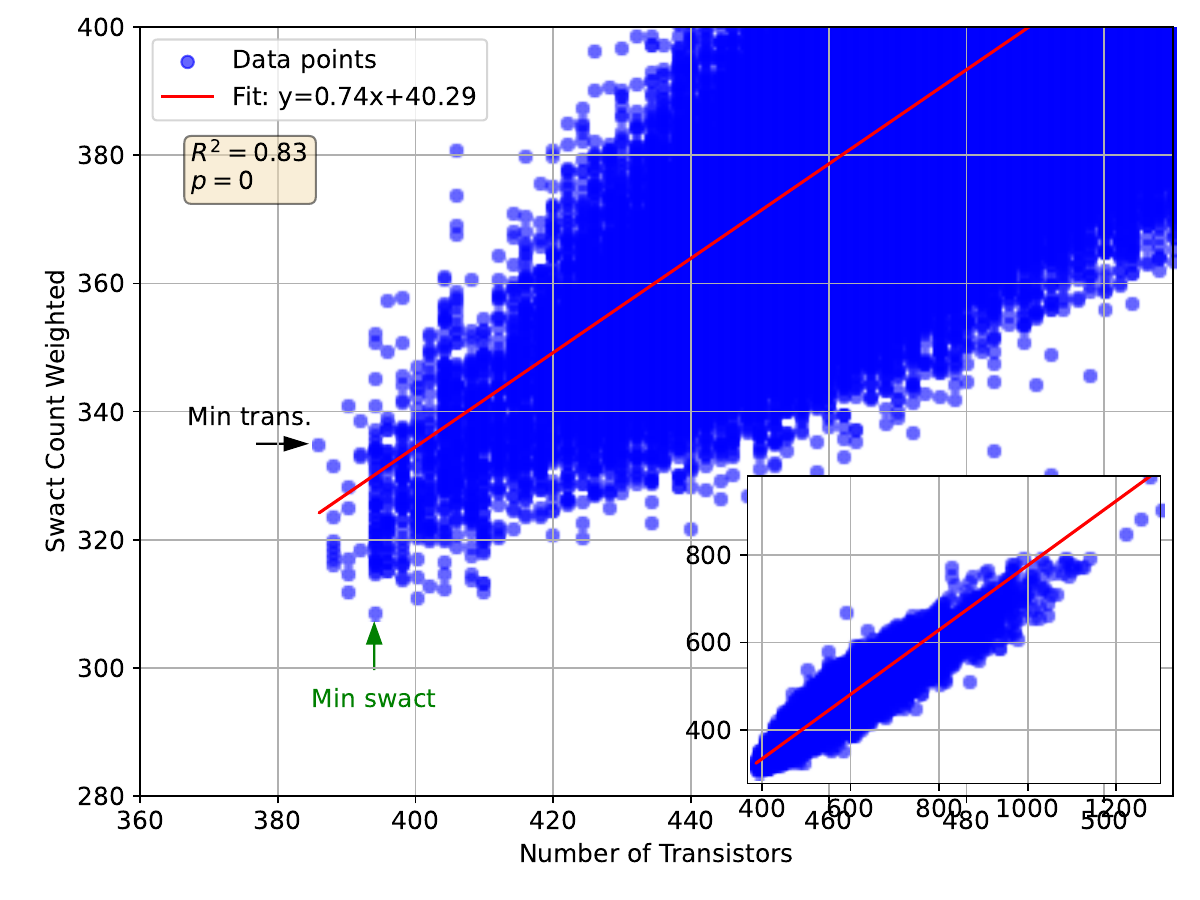}
\label{fig_second_caseC}
}%

\caption{Transistor count and switching activity ($\sigma=3$) of generated circuits for different guided iteration configurations}
\label{fig_sim}
\end{figure}

In the following, we illustrate the performance of the guided random-search-based synthesis framework described in section \ref{sec:optimization}. However, this time we apply different metrics for selection and perform a greater number of steps overall, yielding even more optimized circuits.
As a starting point for the optimization, we define the signed 2\texttimes4-bit multiplier with the star operator in Verilog (\gls{tc}$\rightarrow$\gls{tc} from \textbf{A}).
We run three different experiments and optimize the circuit with different settings: 
a) No iterations (no guided selection): 1 iteration with a chain length of 2000. b) Iterations with target transistor count: 40 iterations each with a chain length 50. c) Iterations with target \gls{swact}: 40 iterations each with a chain length 50.
The total compute effort is equivalent for all experiments with 200 runs each with 2000 steps taken.
Thus, per configuration $2000 \cdot 200 = 400'000$ circuits are produced which are plotted in Figure \ref{fig_sim} with respect to their area and power.
We observe that:
\begin{itemize}
  \item Iterations with guided selection (b and c) improve the overall synthesis \acrshort{qor}, both in terms of area and power.
  \item Selecting for the target metrics transistor count (b) or \gls{swact} (c) produce similar results, but the transistor count target ends up with a slight advantage.
  \item However, choosing the final circuit for the lowest power versus the lowest area (see arrows in Fig. \ref{fig_sim}), the power can be reduced by 5-10\%.
\end{itemize}

The results above led us to conduct synthesis with even higher effort on the multipliers of configurations \textbf{A} and \textbf{B} (described in Table \ref{tab:results_swact}), to validate our observations.
This time, each run contains 50 iterations, each with a chain length of 80, and 5 chains in parallel. 
Each run accounts for a total of $50 \cdot 80 \cdot 5 = 20'000$ steps, that we repeat 200 times for both multipliers.
The resulting circuit \textbf{A} is reduced down to 368 transistors with an average \gls{swact} of 307\,a.u. at $\sigma=3.0$. The multiplier in circuit \textbf{B} is reduced by a modest amount to 370 transistors -- but with an average \gls{swact} of 220\,a.u. at $\sigma=3.0$.
Adding the 2\texttimes36\,a.u. of the low-effort synthesized encoders of Table \ref{tab:results_swact}, yields an overall \gls{swact} of 292\,a.u. for configuration \textbf{B} -- which is still $5\%$ lower than the baseline \textbf{A} without further optimization of the encoder.

\section{Limitations and Outlook}

While our methodology for optimizing multiplier circuits has demonstrated promising improvements in area and power-efficiency, several limitations should be acknowledged:

\begin{enumerate}
    \item \textbf{Bit-width Restriction:} Our experimental evaluation is limited to 4-bit input multipliers without saturation. This constraint simplifies the design space and facilitates rapid exploration, but it may not capture the full complexity or scaling challenges encountered in higher bit-width multipliers. Nevertheless, internal evaluation of configurations \textbf{A} and \textbf{B} with 8-bit input operands tend to indicate that the results are reproducible for larger designs.

    \item \textbf{Fixed-Point Representation Limitation:} Our evaluation exclusively employs fixed-point representations for data encoding and arithmetic operations. While fixed-point arithmetic is widely used for its simplicity and energy efficiency, extending our approach to include floating-point formats could offer additional insights and broaden the applicability of our approach.

    \item \textbf{Simplified Modeling Assumptions:} The \gls{swact} model used in this work weights each cell solely by its transistor count. Although this is effective for comparative studies, it represents a simplification of the dynamic behavior in more complex designs.

    \item \textbf{Synthesis-Only Evaluation:} The experiments presented in this work are restricted to synthesis-level analysis. No place-and-route or timing analysis is performed, so potential issues such as signal race glitching and layout-dependent performance bottlenecks are not captured. Additional tuning during the physical design phase may be required to meet all system-level constraints. 
    
    \item \textbf{Design Space Bias:} Our approach begins from specific starting points (e.g., the multiplier defined by a star operator in Verilog). The initial choice can bias the search towards local optima \cite{gardnerbias}, and additional work is required to understand how these starting points affect the generality of the results.
    
    \item \textbf{Limited Exploration of Data Representations:} Although we compare two's complement, sign-magnitude, and extended sign-magnitude formats, the study is confined to these representations. Other encoding schemes or enhancements (such as handling saturation or overflow differently than what we do here) might be necessary for broader applications.

    \item \textbf{Depth Optimization:} Depth was not a primary focus of our optimization process. However, further synthesis of the combined circuit, including both encoder and multiplier, could help reduce it. Alternatively, if speed is a critical concern, a pipeline stage could be introduced at the encoder output. This approach would require fewer registers compared to traditional methods, thanks to the bottleneck created by the decomposition strategy.
    \item \textbf{Optimization Effort:} Although we put an emphasis on synthesis, in the highest effort scenario (Section \ref{sec:further}) we synthesized only the \gls{tc}$\rightarrow$\gls{tc} and \gls{tc}$\rightarrow$\gls{sme} multipliers. For a comprehensive evaluation, all blocks should be synthesized under this level of effort.

    \item \textbf{Application Regime of Sign-Magnitude:} In this report, we apply sign–magnitude encoding solely to the multiplication stage. However, it would be valuable to explore its effects when extended across the entire data path — for example, on weight storage and activation values.

\end{enumerate}

In future work, it will be crucial to address the limitations of our current methodology. Extending our approach to higher bit-width multipliers will be essential to capture the complexities of larger designs and scaling challenges. Furthermore, exploring automatic methods, such as machine learning, to identify optimal data representations could help generalize our approach beyond the fixed-point formats considered in this study. Finally, incorporating physical design considerations, including place-and-route, power and timing analysis with other tools, would enable a more comprehensive evaluation of the system's performance.

\section{Conclusion}
This work shows that partitioning fixed-point multiplier units into distinct encoder and multiplier components can lead to significant power-efficiency improvements. 
By converting operands from two’s complement to sign-magnitude representation before multiplication, our approach exploits the inherent efficiency of SM encoding to substantially reduce switching activity.
Under a realistic input stream of 4-bit integer values, modeled by a normal distribution with a mean of 0 and a standard deviation of $\sigma=3$, the logic-equivalent transformation achieves up to a 12.9\% reduction in switching activity.
Alternatively, by omitting the representation of the non-symmetric and most negative value (e.g., -8 for 4-bit integers), the resulting logic lowers switching activity by up to 33\%, while keeping the input and output in two’s complement format.
Notably, a multiplication with inputs and outputs entirely in the sign-magnitude domain was shown to reduce switching activity by up to 68.8\%. 
These findings suggest that leveraging explicit data encoding could be a promising strategy to enable synthesis tools to design more power-efficient circuits.
Moreover, adopting modest deviations from the conventional two’s complement encoding can deliver considerable benefits in high-performance computing data flows. 
Consequently, our future work will focus on automating the search for optimal encodings and extending our investigation beyond 4-bit multipliers to assess broader synthesis and architectural potential.

\bibliographystyle{plain}
\bibliography{references}

\end{document}